\documentclass[sn-mathphys,Numbered]{sn-jnl}
\usepackage{manyfoot}
\usepackage{amsmath,amssymb,amsfonts}
\usepackage[ruled,linesnumbered]{algorithm2e}
\usepackage{graphicx}
\usepackage{textcomp}
\usepackage{booktabs}
\usepackage{multirow}
\usepackage{color}
\usepackage{subfig}
\usepackage{overpic}    
\usepackage{hyperref}
\usepackage{cleveref}
\crefname{algocf}{Algorithm}{Algorithms}

\begin{document}

\title[Article Title]{Causal Disentanglement Hidden Markov Model for Fault Diagnosis}

\author[1]{\sur{Rihao Chang}}\email{changrihao@tju.edu.cn}
\equalcont{These authors contributed equally to this work.}

\author[1]{\sur{Yongtao Ma}}\email{mayongtao@tju.edu.cn}
\equalcont{These authors contributed equally to this work.}

\author*[2]{\sur{Weizhi Nie}}\email{weizhinie@tju.edu.cn}

\author[3]{\sur{Jie Nie}}\email{niejie@ouc.edu.cn}
\equalcont{These authors contributed equally to this work.}

\author[2]{\sur{An-an Liu}}\email{anan0422@gmail.com}
\equalcont{These authors contributed equally to this work.}

\affil[1]{\orgdiv{School of Electronics}, \orgname{Tianjin University}, \orgaddress{\street{No.92 Weijin Road}, \city{Tianjin}, \postcode{300072}, \country{China}}}

\affil*[2]{\orgdiv{School of Electrical and Information Engineering}, \orgname{Tianjin University}, \orgaddress{\street{No.92 Weijin Road}, \city{Tianjin}, \postcode{300072}, \country{China}}}

\affil[3]{\orgdiv{Faculty of Information Science and Engineering}, \orgname{Ocean University of China}, \orgaddress{\street{No.238 Songling Road}, \city{Qingdao}, \postcode{266100}, \state{Shandong}, \country{China}}}

\abstract{In modern industries, fault diagnosis has been widely applied with the goal of realizing predictive maintenance. The key issue for the fault diagnosis system is to extract representative characteristics of the fault signal and then accurately predict the fault type. In this paper, we propose a Causal Disentanglement Hidden Markov model (CDHM) to learn the causality in the bearing fault mechanism and thus, capture their characteristics to achieve a more robust representation. Specifically, we make full use of the time-series data and progressively disentangle the vibration signal into fault-relevant and fault-irrelevant factors. The ELBO is reformulated to optimize the learning of the causal disentanglement Markov model. Moreover, to expand the scope of the application, we adopt unsupervised domain adaptation to transfer the learned disentangled representations to other working environments. Experiments were conducted on the CWRU dataset and IMS dataset. Relevant results validate the superiority of the proposed method. }


\maketitle

\section{Introduction}
\label{sec:introduction}
With the rapid development of the social economy and the continuous progress of science and technology, human society has put forward higher requirements for industrial production capacity, and the production equipment of modern industry shows an increasingly complex trend in terms of structure and automation. To ensure the safety of industrial production, the reliability and stability of industrial equipment have become the most important aspects of system design. In other words, the unreliability of the industrial system can lead to a large negative impact on social stability and economic development. Under this condition, fault diagnosis, prediction, and health management have become a key field and gained increasing attention and application\citep{overview, shiqian2020review, yan2022survey, wang2017research}.

Generally, the current fault diagnosis approaches can be divided into two categories: model-based methods and data-driven approaches\citep{tidriri2016bridging}. The model-based methods require a lot of prior knowledge, which makes it difficult to be applied in complex industrial scenarios. The data-driven fault detection method has been proven to be effective for learning fault characteristics based on large-scale data. Principal component analysis (PCA)\citep{PCA}, partial least squares (PLS)\citep{PLS}, canonical correlation analysis (CCA)\citep{CCA}, and independent component analysis (ICA) \citep{ICA} are basic methods of machine learning that have been deeply promoted and studied. However, the aforementioned approaches face challenges for application in complex fault diagnosis conditions. Due to the development of deep learning technology, deep neural networks show their superiority in automatically learning fault representations with more rapid signal processing speed and more robust signal features \citep{b1,b2,b3}. Nevertheless, these deep-learning networks mostly concentrate on the network structure to improve diagnostic performance, which constrains their ability in robustness and generalization.

The bearing vibration signal contains various information and inevitably introduced fault-irrelevant factors for deep learning methods, which reduces the efficiency of feature representations. To this end, we propose the Causal Disentanglement Hidden Markov model (CDHM). First, we propose a causal disentanglement Markov model to make up for the lack of representation ability. By modeling the causal relation of the bearing system, we aim to learn the true causation of the bearing failures. To be specific, we build a sequential network and adopt the learning procedure of Variational Auto-Encoders (VAE) \citep{VAE} to disentangle the input vibration signals into fault-relevant factors and fault-irrelevant factors. The fault-relevant factors contribute to the improvement of diagnostic accuracy because the disentanglement guarantees its information purity and representation effectiveness. 

However, the deep-learning methods trained in specific conditions can hardly be generalized to other work environments, which violates the need for transfer ability in industrial applications. To solve this problem, we introduce domain adaptation into the causal disentanglement hidden Markov model to transfer source disentangling ability into target working conditions. The disentangled representations have an innate advantage in domain alignment tasks. The fault-relevant factor captures the essential cause of the failure while the fault-irrelevant factor represents the working conditions. Therefore, we use Maximum Mean Discrepancy (MMD) \citep{MMD} to align the similar parts from the fault-relevant factors and use the discriminative loss to distinguish differences between fault-irrelevant factors. Note that the transferring is learned in an unsupervised manner. Consequently, CDHM is able to handle the robust representation and the domain adaptation problem. 

The main contributions of this paper are as follows:
\begin{itemize}
\item We propose a Causal Disentanglement Hidden Markov model (CDHM). CDHM disentangles the fault representation into the fault-relevant part and the fault-irrelevant part, which significantly improves the efficiency and robustness of the learned fault representations.
\item We propose to transfer the learned disentanglement ability to other working conditions in an unsupervised manner. We designed the environmental representation and discriminative loss for aligning the fault representations from different domains and thus enabling the adaptation of CDHM to diverse working conditions.
\item We conduct comparative experiments on the Case Western Reserve University bearing dataset (CWRU) and Intelligent Maintenance System dataset (IMS). Significant improvements have been obtained over the state-of-the-art methods. 
\end{itemize}

\section{Related work}

Fault diagnosis is a common research area in industrial inspection, and it plays an important role in many real applications. The purpose of fault diagnosis is to identify the type of equipment fault and facilitates the repair works. Therefore, in the bearing fault diagnosis task, the challenge is to efficiently and accurately learn the relationship between the vibration data and bearing status. We will briefly discuss the evolution of the deep-learning-based fault diagnosis methods and exhibit the advantage of CDHM compared with these previous methods.

The Convolutional Neural Network (CNN) is one of the most effective deep learning networks. Wen et al. \citep{b1} designed a convolutional neural network based on LeNet-5\citep{Lenet}. They propose a method to convert signals into images for training. Jiang et al. \citep{b2} proposed MultiScale Convolutional Neural Networks (MSCNN) in order to achieve fault diagnosis under many conditions. They introduce a coarse-grained layer into the CNN. The MSCNN can effectively learn sufficient features at different time scales from raw vibration signals. Zhang et al. \citep{b3} proposed a few-shot learning method based on the Siamese neural network which learns by exploiting sample pairs of the same or different categories. This method is suitable for fault diagnosis with limited data.

Due to the industrial acquisition of generalization, some methods adopt domain adaptation for the transfer of diagnosis ability between various work environments. Tzeng et al. proposed Deep Domain Confusion (DDC) \citep{DDC}, which uses a CNN structure with an adaption layer and a domain confusion loss to learn semantically meaningful descriptions and domain-specific descriptions. Deep CORAL \citep{DeepCORAL} is proposed by Sun et al.. This method uses CORAL loss \citep{coral} in a deep neural network with a similar structure to DDC, the difference is that the MMD distance is replaced by CORAL. Ganin et al. proposed Domain Adversarial Neural Network (DANN) \citep{DANN}, which utilizes adversarial learning for unsupervised domain adaptation. The network is composed of a feature extractor, a label predictor, and a domain classifier. The feature extractor evolves during the game between the predictor and the classifier. Another efficient method is Deep Convolutional Transfer Learning Network (DCTLN) \citep{DCTLN}. It consists of two modules: conditional recognition and domain adaptation. The condition recognition is implemented by a one-dimensional CNN. The one-dimensional CNN includes a feature extractor and a health condition classifier. Domain adaptation is accomplished by a domain classifier and a distribution difference metric. The domain adaptation module is connected to the feature extractor to help the 1D CNN learn domain invariant features. A recent work of Deep Adversarial Subdomain Adaptation Network (DBSAN) \citep{DASAN} is proposed by Liu et al. DASAN is an unsupervised adversarial domain adaptation method that introduces pseudo-labels for target domains and employs Labeled MMD loss to minimize the distance between category-level subdomains. Besides, this network is also trained in an adversarial manner.

However, these methods neglect the efficiency of feature representation and thus lose the robustness gained from it. To overcome this disadvantage while inheriting previous advantages, we propose CDHM, which introduces the causal disentanglement hidden Markov Model into the fault diagnosis with domain transfer tasks.

\section{Method}
\begin{figure*}[!htp]
  \centering
  \includegraphics[width=0.98\linewidth]{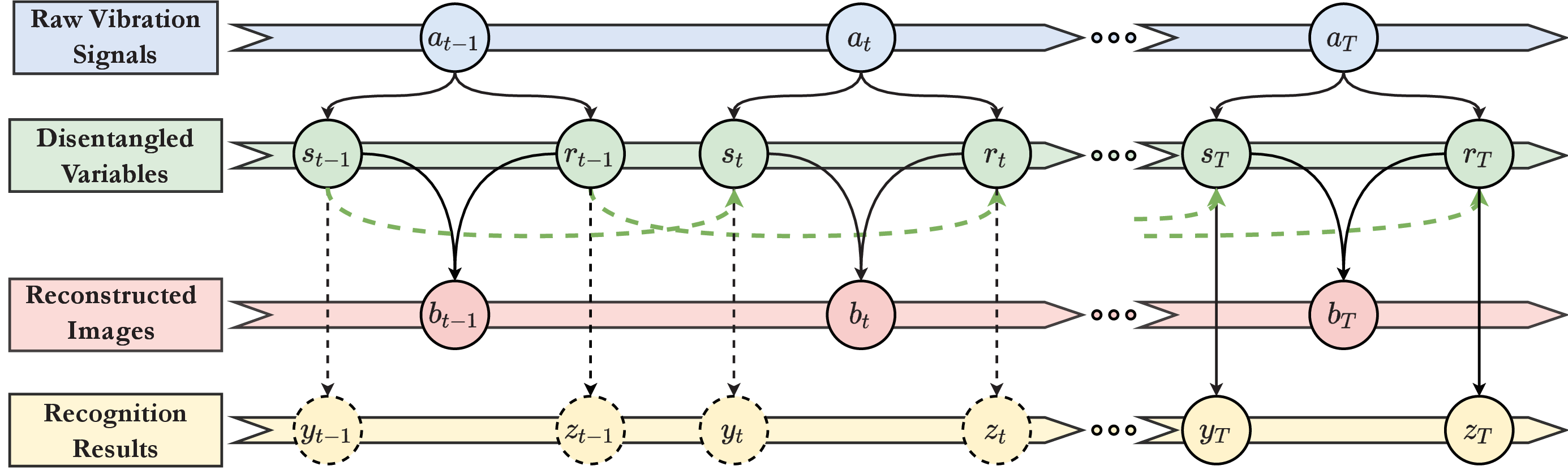}
  \caption{Directed Acyclic Graph of bearing fault diagnosis. The raw vibration signals are the input data. We disentangled the input into causal variables $\boldsymbol{s}$ and $\boldsymbol{r}$. The time-frequency image $\boldsymbol{b}$ is reconstructed from the disentangled variables. The fault type $\boldsymbol{y}$ and domain label $\boldsymbol{z}$ are generated from the learned causal variables. }
  \label{SCM}
\end{figure*}

CDHM structures a time-series causal disentanglement model for bearing fault diagnosis and further extends this model to handle domain transfer problems. We will first introduce the structure of the causal disentanglement model and then present the learning strategy of both causal learning and transfer learning. The setup of our causal disentanglement hidden Markov model is defined in the following paragraph, which serves as the entry of CDHM.\\
\noindent{\textbf{Problem Setup.}} To explore the potential causal mechanism between bearing status and vibration signals, we introduced the causal disentanglement Markov model to describe the fault diagnosis system. Here, the causal relations among variables in the bearing fault diagnosis system are defined as a Directed Acyclic Graph (DAG), as shown in Fig.\ref{SCM}. The variable $\boldsymbol{a}_t \in \boldsymbol{A}$ is the raw vibration data at the time stage $t$. We use sliding window sampling to generate the time-series vibration data. To observe the vibration characteristics from different aspects, we convert the time-series data into time-frequency image $\boldsymbol{b}_t$ by the Continuous Wavelet Transform (CWT) \citep{CWT}. Given total time stages $T$, we have $t \in N^+$ and $t \in \left[1, T\right]$. The $\boldsymbol{y}$ is the predicted fault status. Similarly, $\boldsymbol{z}$ is the predicted data domain split. The disentangled vibration features are defined as the fault-relevant feature $\boldsymbol{s}_t$ and the fault-irrelevant features $\boldsymbol{r}_t$. The $\widetilde{\boldsymbol{b}_t} \in \widetilde{\boldsymbol{B}}$ is the reconstructed observation from the disentangled features. 

\subsection{Causal Disentanglement Hidden Markov Model}
According to the principles of Structured Causal Mechanism\citep{CMC}, the causal relation of variable $\boldsymbol{v}$ in the DAG is formulated as $f_v := \left(pa\left(\boldsymbol{v}\right), \boldsymbol{\epsilon}_v\right)$. Here, $pa\left(\boldsymbol{v}\right)$ denotes the parent variable set of node $\boldsymbol{v}$ and $\boldsymbol{\epsilon}_v$ is the independent exogenous variable, which is determined by system external conditions and not affected by observations. Therefore, we can formulate the correlation of variables in Fig.\ref{SCM} as follows:
\begin{align*}
  &\boldsymbol{s} \rightarrow f_t^s\left(\boldsymbol{a}_t, \boldsymbol{s}_{t-1}, \boldsymbol{\epsilon}_t^s\right), \boldsymbol{r} \rightarrow f_t^r\left(\boldsymbol{a}_t, \boldsymbol{r}_{t-1}, \boldsymbol{\epsilon}_t^r\right),
  \widetilde{\boldsymbol{b}} \rightarrow f_t^{\widetilde{b}}\left(\boldsymbol{s}_t, \boldsymbol{r}_t, \boldsymbol{\epsilon}_t^{\widetilde{b}}\right),\\
  &\boldsymbol{y} \rightarrow f_T^y \left(\boldsymbol{s}_T, \boldsymbol{\epsilon}_T^y\right), \boldsymbol{z} \rightarrow f_T^z\left(\boldsymbol{r}_T, \boldsymbol{\epsilon}_T^z\right).
\end{align*}

Causal Markov Condition (CMC) \citep{CMC} enables the disentanglement of causal variables in the DAG\citep{li2021causal}, according to which, we transform the bearing fault diagnosis task into a Markov time-series analysis task. For the bearing fault diagnosis task, we are eager to strip out the fault-relevant information to improve diagnostic accuracy. According to the basic signal-processing solution of fault diagnosis, the vibration components of the bearing system can be roughly divided into two categories. One is the inherent vibration of the bearing which would alter with the test platform. The second is the vibration caused by bearing faults which exert regular effects on vibration signals. Therefore, we decomposed the raw vibration data $\boldsymbol{a}_t$ into the fault-relevant data $\boldsymbol{s}_t$ and fault-irrelevant data $\boldsymbol{r}_t$. These variables evolve as the intrinsic driver for the progression of the bearing fault status observations. For simplicity, we define $\boldsymbol{m}:=\left\{\boldsymbol{a},\boldsymbol{b}\right\}, \boldsymbol{n}:=\left\{\boldsymbol{r},\boldsymbol{s}\right\}, \boldsymbol{o}:=\left\{\boldsymbol{y},\boldsymbol{z}\right\}$. According to CMC, the joint distribution of the diagnosis model can be factorized as follows:
\begin{align}\label{eq1}
  p\left(\boldsymbol{m}_{t \leq T}, \boldsymbol{n}_{t \leq T}, \boldsymbol{o}_{t=T}\right) = p\left(\boldsymbol{o}_T \mid  \boldsymbol{n}_{T}\right) \ast 
  \prod_{t=1}^{T}\left(p\left(\boldsymbol{n}_t \mid \boldsymbol{n}_{t-1}, \boldsymbol{a}_{t}\right)p\left(\boldsymbol{b}_t \mid \boldsymbol{n}_t\right)\right).
\end{align}

To disentangle the latent variables, we need to analyze their prior distribution $p_{\theta}\left(\boldsymbol{n}_t \mid \boldsymbol{n}_{t-1}, \boldsymbol{a}_{t}\right)$. As the prior distribution is intractable, we use the posterior distribution $q_{\phi}\left(\boldsymbol{n}_t \mid \boldsymbol{m}_t, \boldsymbol{o}_T\right)$ to approximate $p_{\theta}$. Additionally, the terms $p\left(\boldsymbol{o}_T \mid  \boldsymbol{n}_{T-1}\right)$ and $p\left(\boldsymbol{b}_t \mid \boldsymbol{n}_t\right)$ in Eq.\eqref{eq1} are the post-processing of learned causal variables.
Here, we construct a sequential neural network based on VAE to learn the causal disentanglement Markov model. The network architecture is shown in Fig.\ref{Network}. Each unit of the sequential neural network contains a Prior Network and a Posterior Network. 

\subsubsection{Prior Network} 
\begin{figure*}[!htbp]
  \centering
  \includegraphics[width=1\linewidth]{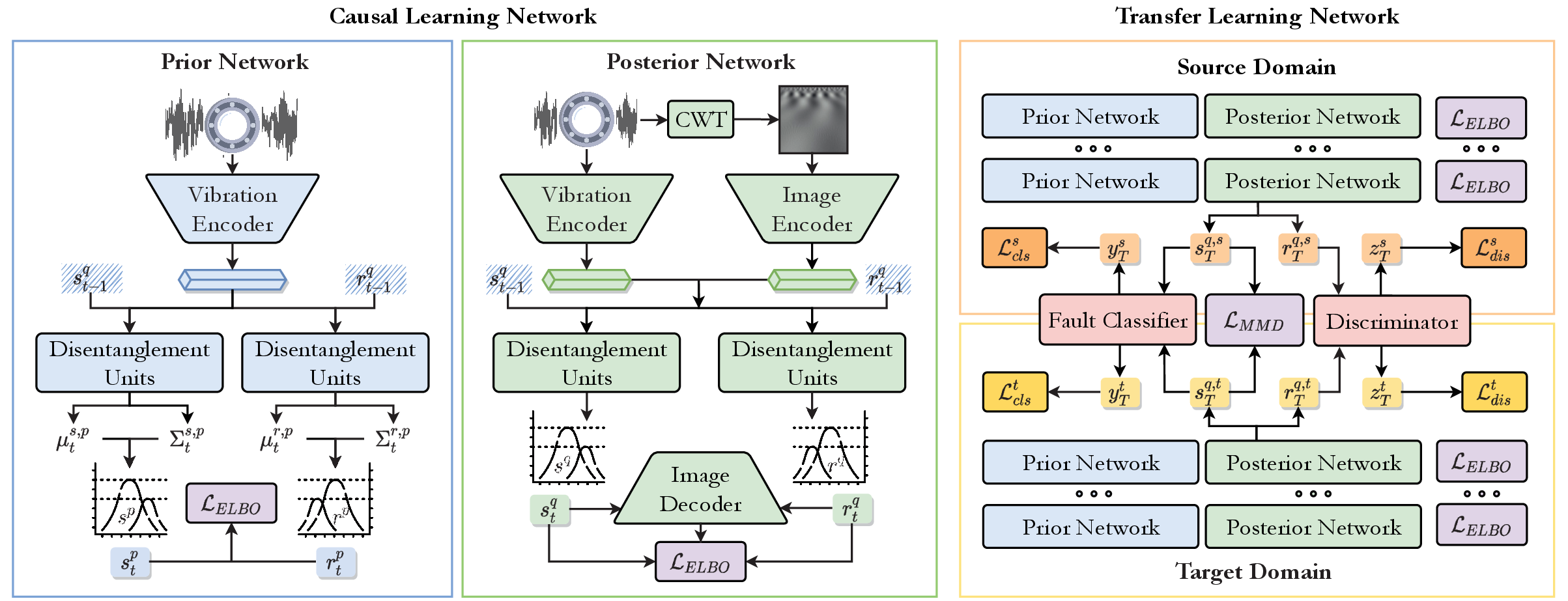}
  \caption{The network architecture of CDHM. CDHM consists of the causal learning network and the transfer learning network. In the causal learning network, we design the pior network and the posterior network to learn the causally disentangled representations $\boldsymbol{s}$ and $\boldsymbol{r}$. The prior network structures the causality to separate the fault-relevant factor $\boldsymbol{s}$ for fault diagnosis. The posterior network additionally considers the observations to approximately solve the prior distribution. The ELBO is reformulated as the learning objective for causal disentanglement. In the transfer learning network, we perform unsupervised domain adaptation on the learned disentangled representations. The MMD loss $\mathcal{L}_{\text{MMD}}$ is used to learn the domain-invariant features for the transfer of causal disentangling ability. The discriminative loss $\mathcal{L}_{\text{dis}}$ contrastively updates the domain-related representations for fault-irrelevant factor $\boldsymbol{r}$. }
  \label{Network}
\end{figure*}

Traditional representation learning introduced inductive bias to perform the downstream tasks, while it is hard to constrain the network to obey the physical rules. Thus, we model a prior network that simulates the informatic mechanism of bearing vibration. As shown in Fig.\ref{Network}, the prior network accepts the time-series vibration data $\boldsymbol{a}_t$ as input and generates the disentangled representations $\boldsymbol{s}^p_t$ and $\boldsymbol{r}^p_t$. Here, $\boldsymbol{s}^p_t$ represents the vibration information related to the faults, and $\boldsymbol{r}^p_t$ encodes the influences irrelevant to the fault, e.g., the bearing type and the rolling system environment. As the vibration data is temporal, we use Gated Recurrent Units (GRUs) \citep{GRU} to construct the disentanglement unit. Thus, we can learn the context from previous time steps. For the optimization, we use Fully Connected layers (FCs) to obtain the mean vector and the log variance vector of these latent variables. 

After extracting the feature of the observation $\boldsymbol{a}_t$, we associate it with the disentangled features from the previous step to generate the current causal components. We formulate the disentanglement unit of variable $\boldsymbol{s}^p_t$ as follows.
\begin{align}
  &\boldsymbol{G}^{p,s}_{in,t} = \text{concat}\left(\text{Encoder}_p\left(\boldsymbol{a}_t\right),\boldsymbol{s}^p_{t-1}\right),\\
  &\boldsymbol{R}^p_t = \sigma\left(\boldsymbol{W}_R \ast \boldsymbol{G}^{s,in}_t+\boldsymbol{bias}_R\right),\label{eq3}\\ 
  &\boldsymbol{Z}^p_t = \sigma\left(\boldsymbol{W}_Z \ast \boldsymbol{G}^{s,in}_t+\boldsymbol{bias}_Z\right),\\
  &\widetilde{\boldsymbol{H}}^p_t = \tanh\left(\boldsymbol{W}_H \left[\boldsymbol{R}^p_t \odot \widetilde{\boldsymbol{H}}^p_{t-1}\right]+\boldsymbol{bias}_H \right),\\
  &\boldsymbol{s}^p_t = \boldsymbol{Z}^p_t \odot \widetilde{\boldsymbol{H}}^p_t + \left(1-\boldsymbol{Z}^p_t\right) \odot \widetilde{\boldsymbol{H}}^p_t. \label{eq6}
\end{align}

Here, $\sigma$ is the sigmoid function, $\boldsymbol{W}_{\star}$ and $\boldsymbol{bias}_{\star}$ are the parameters of the units. The $\boldsymbol{R}^p_t$, and $\boldsymbol{Z}^p_t$ represent the status of the reset gate and update gate, respectively. With the assistance of gates, the present memory $\widetilde{\boldsymbol{H}}^p_t$ considers whether to store the previous information (from the stage $t-1$). Subsequently, the status of variable $\boldsymbol{s}^p_t$ is updated by the update gate and the memory. The learning of $\boldsymbol{r}^p_t$ can be modeled likewise. 

\subsubsection{Posterior Network}
The prior distribution is often empirically assumed to describe the prior beliefs about the distribution of the parameters. From the perspective of Bayesian statistics, the posterior distribution considers the observation of statistic samples and thus enables the updating of the prior distribution. Considering the optimization of VAE, the true posterior distribution is hard to explicitly calculate \citep{VAE}. Therefore, we employ a simpler posterior distribution to approximate the true posterior distribution. 

The posterior network not only accepts the previous system status as the input but also considers the observation $\boldsymbol{b}_t$. To learn the posterior distribution, the disentanglement units are modeled differently. We combined all engaged variables and use fully-connected layers to generate the mean and the log variance for the posterior distribution. Finally, we adopt these variables into specially designed downstream networks in the post-processing part to guide network learning.

The post-processing utilizes the learned variables to produce the desired classification result (classification) and the terms for the network optimization (reconstruction). As the vibration signal includes the intrinsic component and fault component, the causal variable $\boldsymbol{a}_t$ is disentangled into hidden variables $\boldsymbol{s}_t$ and $\boldsymbol{r}_t$. To guarantee the representation ability of learned variables, we use the causal variable $\boldsymbol{s}_t$ to decide the fault type and $\boldsymbol{r}_t$ to recognize the data domain. To avoid the information loss caused by disentanglement, we reconstruct the observation $\boldsymbol{b}_t$ as $\widetilde{\boldsymbol{b}_t}$ from causal variables $\boldsymbol{s}_t$ and $\boldsymbol{r}_t$.

In the next subsection, we will present the learning strategy for causal disentanglement and the transfer of learned knowledge.

\subsection{Learning Strategy}
The learning strategy is used to optimize the the network. According to the practical function, we divide our learning strategy into causal learning and transfer learning. 

\subsubsection{Causal Learning}
To learn the proposed causal disentanglement hidden Markov model, we adopt a sequential VAE, as shown in Fig.\ref{Network}. According to Eq.\eqref{eq1}, we have the prior distribution $p_{\theta}$, which can be specified as:
\begin{align}
  p_{\theta}\left(\boldsymbol{n}_t \mid \boldsymbol{n}_{t-1}, \boldsymbol{a}_{t}\right)=p_{\theta}\left(\boldsymbol{r}_t \mid \boldsymbol{r}_{t-1}, \boldsymbol{a}_{t}\right) \ast p_{\theta}\left(\boldsymbol{s}_t \mid \boldsymbol{s}_{t-1}, \boldsymbol{a}_{t}\right).
\end{align}

As we have previously discussed, the prior network produces the mean vector and the log variance vector. The normalized distribution that $p_{\theta}\left(\boldsymbol{n}_t \mid \boldsymbol{n}_{t-1}, \boldsymbol{a}_t\right)$ obeys is thus formulated as $\mathcal{N}\left(\mu_{\theta}\left(\boldsymbol{n}_{t-1}, \boldsymbol{a}_t\right), \Sigma_{\theta}\left(\boldsymbol{n}_{t-1}, \boldsymbol{a}_t\right)\right)$.
The posterior network learns a tractable distribution to approximate the prior distribution, and is given by the following:
\begin{align}
  q_{\phi}\left(\boldsymbol{n}_{t \leq T} \mid \boldsymbol{m}_{t \leq T}, \boldsymbol{o}_T\right) =&{} \frac{q_{\phi}\left(\boldsymbol{y}_T \mid \boldsymbol{s}_T\right)q_{\phi}\left(\boldsymbol{z}_T \mid \boldsymbol{r}_T\right)}{q_{\phi}\left(\boldsymbol{o}_T\mid \boldsymbol{m}_{t \leq T}\right)} \ast \sum_{t \leq T}q_{\phi}\left(\boldsymbol{n}_t \mid \boldsymbol{n}_{t-1}, \boldsymbol{m}_t\right).
\end{align}

Here, the first term models the predictions at stage $T$ and $q_{\phi}\left(\boldsymbol{n}_t \mid \boldsymbol{n}_{t-1}, \boldsymbol{m}_t\right)$ obeys $\mathcal{N}\left(\mu_{\phi}\left(\boldsymbol{n}_{t-1}, \boldsymbol{m}_t\right), \Sigma_{\phi}\left(\boldsymbol{n}_{t-1}, \boldsymbol{m}_t\right)\right)$ at each time stage.

The optimization target of this structure is the Evidence Lower BOund (ELBO). According to the previous formulations, ELBO is reformulated as:
\begin{align}
  \mathcal{L}_{\text{ELBO}}={}&\mathbb{E}\left[\mathbb{E}_{q_{\phi}}\left[\log\left(\frac{p_{\theta}\left(\boldsymbol{n}_{t \leq T}, \boldsymbol{m}_{t \leq T}, \boldsymbol{o}_T\right)}{q_{\phi}\left(\boldsymbol{n}_{t \leq T} \mid \boldsymbol{m}_{t \leq T}, \boldsymbol{o}_T\right)}\right)\right]\right], \\
  ={}&\mathbb{E}\left[\log\left(q_{\phi}\left(\boldsymbol{o}_T \mid \boldsymbol{m}_{t \leq T}\right)\right) + \sum_{t=1}^{T}\mathcal{L}_{q_{\phi},p_{\theta}}^{t}\right]. 
\end{align}
The likelihood $q_{\phi}\left(\boldsymbol{o}_T \mid \boldsymbol{m}_{t \leq T}\right)$ in the first term can be formulated as: 
\begin{align}
    \int\left(\Pi_{t=1}^{T} q_{\phi}\left(\boldsymbol{n}_{t} \mid \boldsymbol{m}_{t}, \boldsymbol{n}_{t-1}\right)\right) q_{\phi}\left(y_{T} \mid \boldsymbol{s}_{T}\right) d \boldsymbol{n}_{0} \ldots d \boldsymbol{n}_{T}
\end{align}
Here, the terms at each stage are represented as $\mathcal{L}_{q_{\phi},p_{\theta}}^{t}$, which is specified as follows.
\begin{align}
  \mathcal{L}_{q_{\phi},p_{\theta}}^{t} ={}&\mathbb{E}_{q_{\phi}\left(\boldsymbol{n}_{t} \mid \boldsymbol{n}_{t-1}, \boldsymbol{m}_t\right)}\left[\log \left(p_{\theta}\left(\boldsymbol{b}_t \mid \boldsymbol{n}_t\right)\right)\right]-D_{KL}\left[q_{\phi}\left(\boldsymbol{n}_t \mid \boldsymbol{n}_{t-1}, \boldsymbol{m}_t\right), p_{\theta}\left(\boldsymbol{n}_t \mid \boldsymbol{n}_{t-1}, \boldsymbol{a}_t\right)\right]. \label{ELBO}
\end{align}
where the first term is the reconstruction loss, and the second term is the KL divergence of the latent variables. 

Thus, ELBO can be regarded as a lower bound between log evidence $p_{\theta}$ and KL divergence $D_{KL}$. In a nutshell, ELBO is the minimum difference between the true posterior probability and the approximate posterior probability.

\subsubsection{Transfer Learning}
The information on bearing faults varies due to the rotation system. It is difficult to achieve universal deep-learning algorithms for different systems. Traditional deep-learning algorithms require extra supervised training to fit new bearing systems. Differently, we adopt unsupervised domain adaptation to transfer the disentangling ability to new data scenarios. 

The Maximum Mean Discrepancy (MMD) is a popular discrepancy metric approach used in domain adaptation. It measures the distribution discrepancy between the source and target domains. Here, MMD is a measure of the difference between the mean features of two probability distributions, $P$ and $Q$, in a Reproducing Kernel Hilbert Space (RKHS) with a characteristic kernel $k$. The MMD is formally defined in reference \citep{MMD}. Here, we use $L_{\widehat{\text{MMD}}}$ as an unbiased empirical estimate of $L_{\text{MMD}}$:
\begin{align}
\mathcal{L}_{\widehat{\text{MMD}}}(X_S, X_T) &=  \left\|\frac{1}{N_{s}} \sum_{i=1}^{N_{s}} \psi\left(x_{i}\right)-\frac{1}{N_{t}} \sum_{j=1}^{N_{t}} \psi\left(x_{j}\right)\right\|_{\mathcal{H}}^{2} \\
 &= \frac{1}{N_{s}^{2}} \sum_{i=1}^{N_{s}} \sum_{j=1}^{N_{s}} k\left(x_{i}^{s}, x_{j}^{s}\right)+\frac{1}{N_{t}^{2}} \sum_{i=1}^{N_{t}} \sum_{j=1}^{N_{t}} k\left(x_{i}^{t}, x_{j}^{t}\right) -\frac{2}{N_{s} N_{t}} \sum_{i=1}^{N_{s}} \sum_{j=1}^{N_{t}} k\left(x_{i}^{s}, x_{j}^{t}\right),
\end{align}
where $N_s$ and $N_t$ are the number of samples from each domain. $x^{t} \in X_T$ and $x^{s} \in X_S$ are the random variables sampled from distribution $P$ and $Q$. The RKHS is denoted by $\mathcal{H}$ and $\psi$ is associated with characteristic kernel $k$, which is an inner product between feature mappings, $k(x_s,x_t)=〈\psi\left(x_s\right),\psi\left(x_t\right)〉$.  By learning this function, the consistent cause of bearing faults will be obtained.

According to the previous conclusion, the vibration data can be divided into the fault-relevant variable $\boldsymbol{s}$ and fault-irrelevant variable $\boldsymbol{r}$. Variable $\boldsymbol{s}$ through a classifier gives the predicted fault type. Differently, variable $\boldsymbol{r}$ represents the characteristics of the rotation system, which is exactly the distinguishable information of different data domains. Therefore, we utilize a domain discriminator to learn the domain label $\boldsymbol{z}$ from $\boldsymbol{r}$. Consequently, we can distinguish data between the source and target domains. The discriminative loss is adopted to guide this process.
\begin{align}
  \mathcal{L}_{\text{dis}} = -\frac{1}{M} \sum_{i=1}^{M}\left(\boldsymbol{z}_{i,sr} \log(p_{i,sr}) + \boldsymbol{z}_{i,tr} \log(p_{i,tr})\right),
\end{align}
where $M$ is the sample count, $\boldsymbol{z}$ is the true domain label of current sample, and $p_{i,domain}$ represents the predicted probability of sample $i$ belongs to $domain$. Note that the discriminator is shared between domains. Previously, the fault-irrelevant factor $r$ learns the environmental features. Currently, benefiting from $\mathcal{L}_{\text{dis}}$, factor $\boldsymbol{r}$ contrastively learns the salient domain characteristics.

To verify the fault type diagnostic accuracy, we use the cross-entropy loss to perform the classification, which is formulated as:
\begin{align}
  \mathcal{L}_{\text{cls}} = -\frac{1}{M} \sum_{i=1}^{M} \sum_{c=1}^{C} y_{i,c} \log(p_{i,c}),
\end{align}
where $M$ is the number of samples, $C$ is the number of classes, $y_{i,c}$ is the true label for sample $i$ and class $c$, and $p_{i,c}$ is the predicted probability for sample $i$ and class $c$.

Under these conditions, the overall learning strategy is based on the loss function $\mathcal{L}$:
\begin{align}
  \mathcal{L} = \mathcal{L}_{\text{ELBO}} + \alpha\mathcal{L}_{\widehat{\text{MMD}}} + \beta\mathcal{L}_{\text{dis}} + \gamma\mathcal{L}_{\text{cls}}, \label{lossall}
\end{align}
where $\alpha$, $\beta$, $\gamma$ are the balancing parameters, we experimentally set them as $0.5$, $0.2$ and $0.3$.

\subsection{Workflow}
The workflow of CDHM is shown in the \Cref{Algorithm}. The mainstream following the time series achieves the disentanglement of the causal factors. The source domain and target domain are trained jointly while no fault-type labels are acquired in the target domain. Finally, the network parameters are updated at each epoch by the combined loss function according to Eq.\eqref{lossall}. 

\begin{algorithm}
\RestyleAlgo{ruled}
\SetKwInput{KwNetwork}{Network}
\caption{Working pipeline of CDHM.}\label{Algorithm}
\KwData{Vibration signal: $D^s_v=\{\left(\boldsymbol{v}^s, \boldsymbol{y}^s\right)_i\}^{N_s}_{i=1}$, $D^t_v=\{\left(\boldsymbol{v}^t\right)_j\}^{N_t}_{j=1}$. Converted image: $D^s_i=\{\left(\boldsymbol{b}^s\right)_i\}^{N_s}_{i=1}$, $D^t_i=\{\left(\boldsymbol{b}^t\right)_j\}^{N_t}_{j=1}$.}
\KwNetwork{Vibration encoder: $\text{E}_V$, disentanglement unit: $\text{E}_{D_{\star}}$, image encoder: $\text{E}_I$, Posterior encoder: $\text{F}_{\star}$, image decoder: $\text{D}_I$, classifier: $\text{CLS}$, discriminator: $\text{DIS}$.}
\KwResult{Diagnosed fault type $\widetilde{\boldsymbol{y}}$.}
\tcc{Training starts with the parameter initialization.}
\For{epoch ($e$)}{
    \tcc{Source domain procedure}
    \For{time-step ($t$)}{
        $\boldsymbol{a}_t = \text{E}_V\left(\boldsymbol{v}_t\right)$, $\boldsymbol{s}^p_t = \text{E}_{D_s}\left(\text{concat}\left(\boldsymbol{a}_t, \boldsymbol{s}^q_{t-1}\right)\right)$, $\boldsymbol{r}^p_t = \text{E}_{D_r}\left(\text{concat}\left(\boldsymbol{a}_t, \boldsymbol{r}^q_{t-1}\right)\right)$\;
        $\boldsymbol{u}_t = \text{E}_I\left(\boldsymbol{b}_t\right)$, $\boldsymbol{s}^q_t = \text{F}_s\left(\text{concat}\left(\boldsymbol{a}_t, \boldsymbol{u}_t, \boldsymbol{s}^q_{t-1}\right)\right)$, $\boldsymbol{r}^p_t = \text{F}_r\left(\text{concat}\left(\boldsymbol{a}_t, \boldsymbol{u}_t, \boldsymbol{r}^q_{t-1}\right)\right)$\;
        \tcc{Reparameterization is included in the above encoders.}
        $\widetilde{\boldsymbol{b}}_t = \text{D}_I\left(\boldsymbol{s}^q_t, \boldsymbol{r}^q_t\right)$, $\mathcal{L}_{\text{ELBO}}\left(\boldsymbol{s}^p_t, \boldsymbol{s}^q_t, \boldsymbol{r}^p_t, \boldsymbol{r}^q_t, \boldsymbol{b}_t, \widetilde{\boldsymbol{b}}_t\right)$\;
    }
    $\widetilde{\boldsymbol{y}}_T = \text{CLS}\left(\boldsymbol{s}^q_T\right)$, $\widetilde{\boldsymbol{z}}_T = \text{DIS}\left(\boldsymbol{r}^q_T\right)$\;
    $\mathcal{L}_{\text{cls}}\left(\widetilde{\boldsymbol{y}}_T, \boldsymbol{y}_T\right)$, $\mathcal{L}_{\text{dis}}\left(\widetilde{\boldsymbol{z}}_T, \boldsymbol{z}_T\right)$\;
    \tcc{Target domain follows the same procedure and then calculates the MMD loss.}
    $\mathcal{L}_{\hat{\text{MMD}}}\left(\boldsymbol{s}^{q, src}_T, \boldsymbol{s}^{q, tgt}_T\right)$\;
    \tcc{Update parameters of the network}
}
\emph{Given vibration signal $\boldsymbol{v}$, the testing procedure uses disentangled variable $\boldsymbol{s}_T$ to diagnose the fault type $\widetilde{\boldsymbol{y}}$}\;
\end{algorithm}

\subsection{Network Structure}
CDHM is composed of five modules. For the processing of converted time-frequency images, the image encoder and image decoder are designed based on the 2D convolution layers and 2D transposed convolution layers. The prior encoder utilizes GRU cells to learn the time-series dependency and generate the prior distribution of the desired causal factors. Accordingly, the posterior distribution is encoded by the fully connected layers. The fault classifier and domain discriminator are linear layers with softmax and sigmoid functions, respectively. Finally, we calculated the overall loss function and update the parameters of CDHM.

\begin{sidewaystable*}[hbp]
    \label{netdet}
    \centering
    \caption{The network structure of CDHM.}
     \setlength{\tabcolsep}{1mm}
     \scalebox{0.8}{
        \begin{tabular}{clllcll}
            \toprule
            Network                          & Layer                            & Setting                                              &  & Network                                  & Layer                           & Setting                                \\ \midrule
            \multirow{6}{*}{Image Encoder}   & LinearLayer                      & Linear(signal\_size, 128)                            &  & \multirow{7}{*}{Prior Encoder}           & GRU\_Layer\_s                   & GRUCell(input\_size,256)               \\ \cmidrule{2-3} \cmidrule{6-7} 
                                             & \multirow{3}{*}{ConvLayer * 5}   & Conv2d(kernel\_size=4, stride=2, padding=1)          &  &                                          & \multirow{2}{*}{LinearLayer\_s} & For mul: Linear(input\_size, 128)      \\ \cmidrule{3-3} \cmidrule{7-7} 
                                             &                                  & BatchNorm2d(input\_size)                             &  &                                          &                                 & For logvar: Linear(input\_size,   128) \\ \cmidrule{3-3} \cmidrule{6-7} 
                                             &                                  & ReLU(input\_size)                                    &  &                                          & GRU\_Layer\_r                   & GRUCell(input\_size,256)               \\ \cmidrule{2-3} \cmidrule{6-7} 
                                             & PoolingLayer                     & Adaptive\_avg\_pool2d($1\times1$)                    &  &                                          & \multirow{2}{*}{LinearLayer\_r} & For mul: Linear(input\_size, 128)      \\ \cmidrule{1-3} \cmidrule{7-7} 
            \multirow{7}{*}{Image Decoder}   & \multirow{2}{*}{LinearLayer}     & Linear(128, feature\_map\_dim)                       &  &                                          &                                 & For logvar: Linear(input\_size,   128) \\ \cmidrule{3-3} \cmidrule{5-7} 
                                             &                                  & Leaky\_ReLU(input\_size, 0.2)                        &  & \multirow{7}{*}{Post Encoder}            & \multirow{3}{*}{LinearLayer\_s} & For feature:   Linear(input\_size,128) \\ \cmidrule{2-3} \cmidrule{7-7} 
                                             & \multirow{3}{*}{DeconvLayer * 5} & ConvTranspose2d(kernel\_size=4, stride=2, padding=1) &  &                                          &                                 & For mul: Linear(input\_size, 128)      \\ \cmidrule{3-3} \cmidrule{7-7} 
                                             &                                  & BatchNorm2d(input\_size)                             &  &                                          &                                 & For logvar: Linear(input\_size,   128) \\ \cmidrule{3-3} \cmidrule{6-7} 
                                             &                                  & Leaky\_ReLU(input\_size,0.2)                         &  &                                          & \multirow{3}{*}{LinearLayer\_r} & For feature:   Linear(input\_size,128) \\ \cmidrule{2-3} \cmidrule{7-7} 
                                             & Activation                       & Tanh(input\_size)                                    &  &                                          &                                 & For mul: Linear(input\_size, 128)      \\ \cmidrule{1-3} \cmidrule{7-7} 
            \multirow{5}{*}{Fault Classifer} & \multirow{3}{*}{LinearLayer\_c}  & Linear(256, 128)                                     &  &                                          &                                 & For logvar: Linear(input\_size,   128) \\ \cmidrule{3-3} \cmidrule{5-7} 
                                             &                                  & ReLU(input\_size)                                    &  & \multirow{5}{*}{Discriminator}           & \multirow{3}{*}{LinearLayer\_d} & Linear(256, 128)                       \\ \cmidrule{3-3} \cmidrule{7-7} 
                                             &                                  & Linear(128, class\_number)                           &  &                                          &                                 & ReLU(input\_size)                      \\ \cmidrule{2-3} \cmidrule{7-7} 
                                             & Activation                       & Softmax(input\_size)                                 &  &                                          &                                 & Linear(128, 2)                         \\ \cmidrule{1-3} \cmidrule{6-7} 
            \multicolumn{3}{c}{-}                                                                                                      &  &                                          & Activation                      & Sigmoid(input\_size)                   \\
            \bottomrule
        \end{tabular}
    }
\end{sidewaystable*}

\section{Experiments}
In this section, we will introduce the experimental evaluations from two subsections: the comparisons with domain adaptation methods and the experimental analysis. The analysis includes two experiments: the ablation study and the robustness analysis. 
Before presenting the experimental results, we first introduce our data conditions and implemented details. 

\subsection{Data Descriptions}
To evaluate the disentanglement ability and transfer ability of CDHM, we conduct several experiments on the Case Western Reserve University dataset \citep{CWRU} and Intelligent Maintenance Systems dataset \citep{IMS}. These two datasets all collect the vibration signals of rolling bearings.

\subsubsection{Case Western Reserve University Bearing Dataset}
The Case Western Reserve University (CWRU) bearing dataset is a benchmark dataset that contains vibration signals of normal and faulty ball bearings under different load and speed conditions. 

The experiments are from the drive end bearing with a sampling frequency of 12 kHz. There are four different working loads corresponding to four different speeds. The fault according to the fault location includes the inner race, ball, and outer race. The fault diameters of the three faults, which can represent different severity in the rolling bearing, are 0.007, 0.014, and 0.021 inches, respectively. 

\subsubsection{Intelligent Maintenance Systems Dataset}
The Intelligent Maintenance Systems (IMS) dataset is a collection of bearing acceleration data from three run-to-failure experiments on a loaded shaft. The IMS dataset contains four types of health conditions, including normal, inner race fault, ball fault, and outer race fault. The test conditions are a shaft speed of 2000 rpm and a radial load of 6000 lbs. The data acquisition interval is 10 min, the sampling frequency is 20 kHz, and the number of data points in each file is 20,480.

\subsection{Implemented Details}
In these experiments, the vibration data length of each sample is 1200. We use the sliding window sampling with a step size of 400 and an overlapping of 50\% step size to generate the time-series data for CDHM. Thus, the number of time stages, $T$, is 5 in our experiments. For each dataset, the data is split into training and testing sets at a ratio of 7:3.
The vibration data in each time stage is pre-processed by CWT and generates a time-frequency image at a resolution of $224\times224$. The pretrained ResNet-18 \citep{Resnet} is adopted to learn the features from the CWT images. The one-dimensional convolution layers are used for the learning of the vibration signals.
We use Adam optimizer \citep{Adam} and set the learning rate as $10^{-5}$. We trained the network at a batch size of 32 with 200 iterations.
We developed the evaluations with Python and Pytorch on an Nvidia RTX 3090 GPU, and Matlab is used to pre-process the vibration signals.

\subsection{Comparisons with Domain Adaptation Methods}
The transfer performance is evaluated from two experiments: the transfer task between different workloads and different machines.  Here, we use five methods for the comparisons, including BASE, DDC, Deep CORAL, DANN, and DCTLN. Especially, the baseline (BASE) keeps two convolutional feature extractors for the vibration signal and the converted image. The classifier for fault diagnosis is also included. This structure adopts no adaption algorithm and is directly tested on the target domain.

The comparison results between these methods and our CDHM will be detailed in the following parts. Note that the evaluation metric is mean classification accuracy. 

\subsubsection{Case 1 - Transfer between different working loads}
\begin{figure}[htbp]
	\centering
	\subfloat[Confusion matrix of BASE;]{\includegraphics[width=.3\columnwidth]{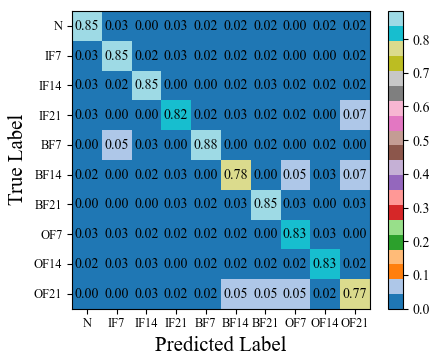}}\hspace{5pt}
	\subfloat[Confusion matrix of DCTLN;]{\includegraphics[width=.3\columnwidth]{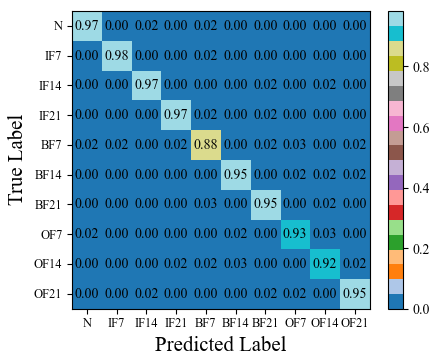}}
	\subfloat[Confusion matrix of CDHM.]{\includegraphics[width=.3\columnwidth]{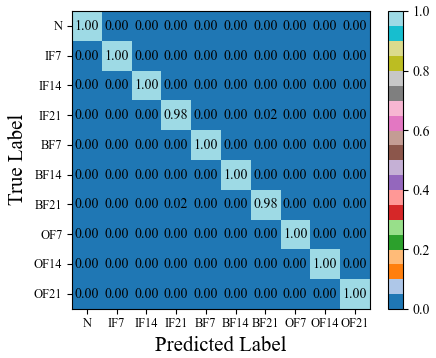}}\hspace{5pt}
	\caption{Visualized results. (a), (b), and (c) are the confusion matrices of BASE, DCTLN and CDHM respectively.}
 \label{analysis}
\end{figure}

In this experiment, we focus on the transfer ability between different bearing working loads from the CWRU dataset. Here, the class is divided by both fault type and working load, and we select 300 samples for each class. The network is trained on the labeled source data and unlabeled target data. We use "source $\rightarrow$ target" to represent the transfer from source data to target data. There are a total of twelve sets of experiments between different workloads, and bidirectional experiments are conducted between each two working loads, e.g. 1hp$\rightarrow$2hp and 2hp$\rightarrow$1hp. The experimental results of the proposed CDHM and five other fault diagnosis methods are shown in Tab.\ref{load}. The best results are displayed in bold. Besides, we also show the comparison results of some confusion matrices of 3hp$\rightarrow$0hp experiments in Fig.\ref{analysis}(a), (b), and (c).
\begin{table*}[ht]
  \centering
  \caption{The performance of the domain adaptation on different working loads.}
  \scalebox{0.75}{
  \begin{tabular}{lllllll}
    \toprule
    Transfer task & \multicolumn{6}{c}{Fault diagnostic accuracy (\%)}                                  \\
    \midrule
    Source→Target       & BASE            & DDC            & Deep CORAL     & DANN           & DCTLN          & CDHM     \\
    \midrule
    0hp → 1hp           & 96.53 ± 0.12    & 95.71 ± 0.41   & 94.52 ± 0.51   & 96.83 ± 0.32   & 98.69 ± 0.06   & \textbf{99.73 ± 0.07} \\
    0hp → 2hp           & 92.84 ± 0.49    & 95.48 ± 0.25   & 92.34 ± 1.10   & 96.24 ± 0.53   & 99.87 ± 0.03   & \textbf{100.0 ± 0.00} \\
    0hp → 3hp           & 88.91 ± 0.93    & 92.19 ± 0.58   & 91.56 ± 2.08   & 97.82 ± 0.36   & 98.92 ± 0.24   & \textbf{99.69 ± 0.09} \\
    1hp → 0hp           & 98.82 ± 0.07    & 98.96 ± 0.43   & 98.33 ± 0.24   & 98.65 ± 0.20   & 100.0 ± 0.00   & \textbf{100.0 ± 0.00} \\
    1hp → 2hp           & 98.56 ± 0.21    & 97.79 ± 0.26   & 98.82 ± 0.09   & 99.37 ± 0.18   & 99.72 ± 0.02   & \textbf{100.0 ± 0.00} \\
    1hp → 3hp           & 92.67 ± 0.18    & 93.48 ± 0.91   & 94.72 ± 1.61   & 99.15 ± 0.32   & 98.93 ± 0.48   & \textbf{99.90 ± 0.02} \\
    2hp → 0hp           & 96.31 ± 0.42    & 96.99 ± 0.72   & 97.53 ± 0.14   & 92.85 ± 0.71   & 99.32 ± 0.18   & \textbf{100.0 ± 0.00} \\
    2hp → 1hp           & 97.15 ± 0.29    & 98.51 ± 0.44   & 99.49 ± 0.11   & 95.47 ± 0.93   & 99.20 ± 0.13   & \textbf{99.92 ± 0.03} \\
    2hp → 3hp           & 98.88 ± 0.06    & 99.22 ± 0.34   & 98.28 ± 0.07   & 99.64 ± 0.09   & 98.85 ± 0.09   & \textbf{100.0 ± 0.00} \\
    3hp → 0hp           & 81.09 ± 1.13    & 84.35 ± 0.81   & 88.37 ± 0.13   & 87.12 ± 0.93   & 93.03 ± 0.10   & \textbf{98.72 ± 0.18} \\
    3hp → 1hp           & 84.56 ± 0.75    & 89.61 ± 0.74   & 90.43 ± 0.38   & 87.48 ± 0.73   & 92.53 ± 0.76   & \textbf{98.04 ± 0.21} \\
    3hp → 2hp           & 95.93 ± 0.53    & 96.49 ± 0.63   & 97.03 ± 0.16   & 97.42 ± 0.16   & 99.41 ± 0.12   & \textbf{100.0 ± 0.00} \\
    Average             & 93.52           & 94.90          & 95.12          & 95.67          & 98.21          & \textbf{99.67}        \\
    \bottomrule 
  \end{tabular}
  }
\label{load}
\end{table*}

The vibration varies with the working loads because the loads define the rotating speed of the bearings. Accordingly, diagnosing bearing faults from different working loads will greatly reduce repetitive operations in industrial practice. In Tab.\ref{load} and Fig.\ref{analysis}, CDHM achieves the best accuracy among the comparison methods, but the performance gap between these candidates is not significant. It's because the data from different working loads shares sufficient relevant data for fault diagnosis methods, while the improvement from the causal disentanglement is not so obvious as in the following case.

\subsubsection{Case 2 - Transfer between different machines}
Since the transfer from different loads still maintains the majority of the vibration characteristics unchanged, we conduct experiments on different machines to verify the effectiveness of the causal disentanglement. Here, we choose 1000 samples for each fault type from both CWRU and IMS. Note that data with the smallest fault size of CWRU are used due to IMS is a run-to-failure dataset.

Considering the data environments are usually different while still related, the network should be able to recognize the faults from these different machines. To demonstrate this generalization performance of CDHM, we evaluated it on the transfer tasks between CWRU and IMS. The transfer results are shown in Tab.\ref{machine}. The best results are displayed in bold.
\begin{table*}[htbp]
  \centering
  \caption{The performance of the domain adaptation on different machines.}
  \scalebox{0.74}{
  \begin{tabular}{lllllll}
  \toprule
  \multirow{2}{*}{Transfer task}  & \multicolumn{6}{c}{Fault diagnostic accuracy (\%)}            \\
  \cmidrule(lr){2-7}
                                  &  BASE          &  DDC           &  Deep CORAL    &  DANN          &  DCTLN         &  CDHM     \\
  \midrule
  CWRU→IMS                        &  68.75 ± 2.79  &  75.95 ± 1.13  &  77.25 ± 1.02  &  79.80 ± 1.47  &  80.72 ± 0.74  &  \textbf{89.65 ± 0.33} \\
  IMS→CWRU                        &  54.32 ± 2.13  &  59.77 ± 2.16  &  64.89 ± 1.71  &  68.21 ± 1.53  &  72.03 ± 1.29  &  \textbf{80.24 ± 0.58} \\
  Average                         &  61.54         &  67.86         &  71.07         &  74.01         &  76.38         &  \textbf{84.95}      \\
  \bottomrule
  \end{tabular}
  }
\label{machine}
\end{table*}

According to Tab.\ref{machine}, CDHM significantly outperforms the other methods on the transfer between different machines. Considering the results in Case \uppercase\expandafter{\romannumeral1}, the improvement here is largely benefited from the causal disentanglement which captures the natural intrinsic of bearing faults and excludes the influence from the irrelevant causal factors.

\subsubsection{Discussion}
According to the results in the aforementioned cases, CDHM realizes the best accuracy and our baseline (BASE) is the performance lower bound. This is because BASE has no task-specific design except for the multi-modal features being directly concatenated. The DDC and Deep CORAL both focus on the domain alignment between source data and target data using MMD-like loss functions. In CDHM, not only MMD loss is adopted to capture the data relevance, but it also utilizes the discriminative domain-specific features to refine the adaptation of cross-domain data in a contrastive way. DANN utilized adversarial learning for domain adaptation, while its performance is still not as satisfying as CDHM. Moreover, DCTLN introduced domain labels for domain adaptation. This strategy innately suits our disentangled fault-irrelevant features. It is because the fault-irrelevant features physically represent the characteristics of the bearing system which mainly contributes to the divergence of different bearing data domains. As a conclusion, CDHM not only disentangles the fault features through causal learning but also adapts transfer learning to the decoupled features and finally achieves the best fault diagnostic accuracy. 

\subsection{Analysis}
\subsubsection{Analysis on loss functions}

CDHM adopts causal learning and transfer learning strategies to achieve the aforementioned performance. To verify the effectiveness of these learning strategies, we conduct the ablation study on the loss functions. In this experiment, we choose the transfer experiment from CWRU to IMS, and three loss variants named $\mathcal{L}_1$ to $\mathcal{L}_3$ are listed in Tab.\ref{abloss}.

For a pure comparison, we remove all the loss functions except for the necessary $\mathcal{L}_{\text{cls}}$ to form $\mathcal{L}_1$. The change made by $\mathcal{L}_1$ requires a comparable change of the network. Here, we use the previously mentioned baseline, BASE, for $\mathcal{L}_1$. Furthermore, we individually evaluated the performance of the causal learning and transfer learning based on the CDHM network. For transfer learning, we remove $\mathcal{L}_{\text{ELBO}}$ and $\mathcal{L}_{\text{dis}}$ to establish a pure transfer learning network trained with a loss $\mathcal{L}_2$. On the contrary, $\mathcal{L}_{\text{ELBO}}$ and $\mathcal{L}_{\text{cls}}$ are preserved, which resulted in $\mathcal{L}_3$.  The experimental results are listed in Tab.\ref{abloss}.

\begin{table*}[htbp]
  \centering
  \caption{The experimental results of the ablation study on the loss functions.}
    \scalebox{1}{
      \begin{tabular}{lcc}
      \toprule
      Loss function                                                                                                                      & Accuracy  & Symbol                                   \\
      \midrule
      $\mathcal{L}_{\text{cls}}$                                                                                         & 68.75 ± 2.79     & $\mathcal{L}_1$                                      \\
      $\mathcal{L}_{\widehat{\text{MMD}}} \ \& \ \mathcal{L}_{\text{cls}}$                                 & 78.68 ± 1.36     & $\mathcal{L}_2$                            \\
      $\mathcal{L}_{\text{ELBO}} \ \& \ \mathcal{L}_{\text{cls}}$                                                             & 76.89 ± 1.15     & $\mathcal{L}_3$                               \\
      $\mathcal{L}_{\text{ELBO}} \ \& \ \mathcal{L}_{\widehat{\text{MMD}}} \ \& \ \mathcal{L}_{\text{dis}} \ \& \ \mathcal{L}_{\text{cls}}$         & 89.65 ± 0.33     & $\mathcal{L}$                                        \\
      \bottomrule
      \end{tabular} 
  }
\label{abloss}
\end{table*}

The experimental results above reflect the rationale of the learning strategy. 
The removal of all specific-designed loss functions except for $\mathcal{L}_{\text{cls}}$ makes $\mathcal{L}_1$ a simplified version of the other loss functions and inevitably achieves undesirable results. The purpose of using $\mathcal{L}_1$ is to have a baseline for comparison to see the impact of each specific-designed learning strategy. 
The individual evaluations of causal learning and transfer learning have shown satisfying accuracies. Although on the transfer tasks, $\mathcal{L}_2$ with specially designed domain adaptation strategy achieves minorly better results than the casual learning variant $\mathcal{L}_3$, this reflects the disentanglement ability of causal network could capture the intrinsic cause of the failure and weaken the influence of the domain-drift.
Afterward, the combination of each loss function results in the final optimization target $\mathcal{L}$ and realizes the best fault diagnostic accuracy. These experiments indicate that the learning strategy of CDHM is efficient in bearing fault diagnosis tasks.

\subsubsection{Robustness under noisy environments}
\begin{figure}[!htbp]
  \centering
  \includegraphics[width=0.5\linewidth]{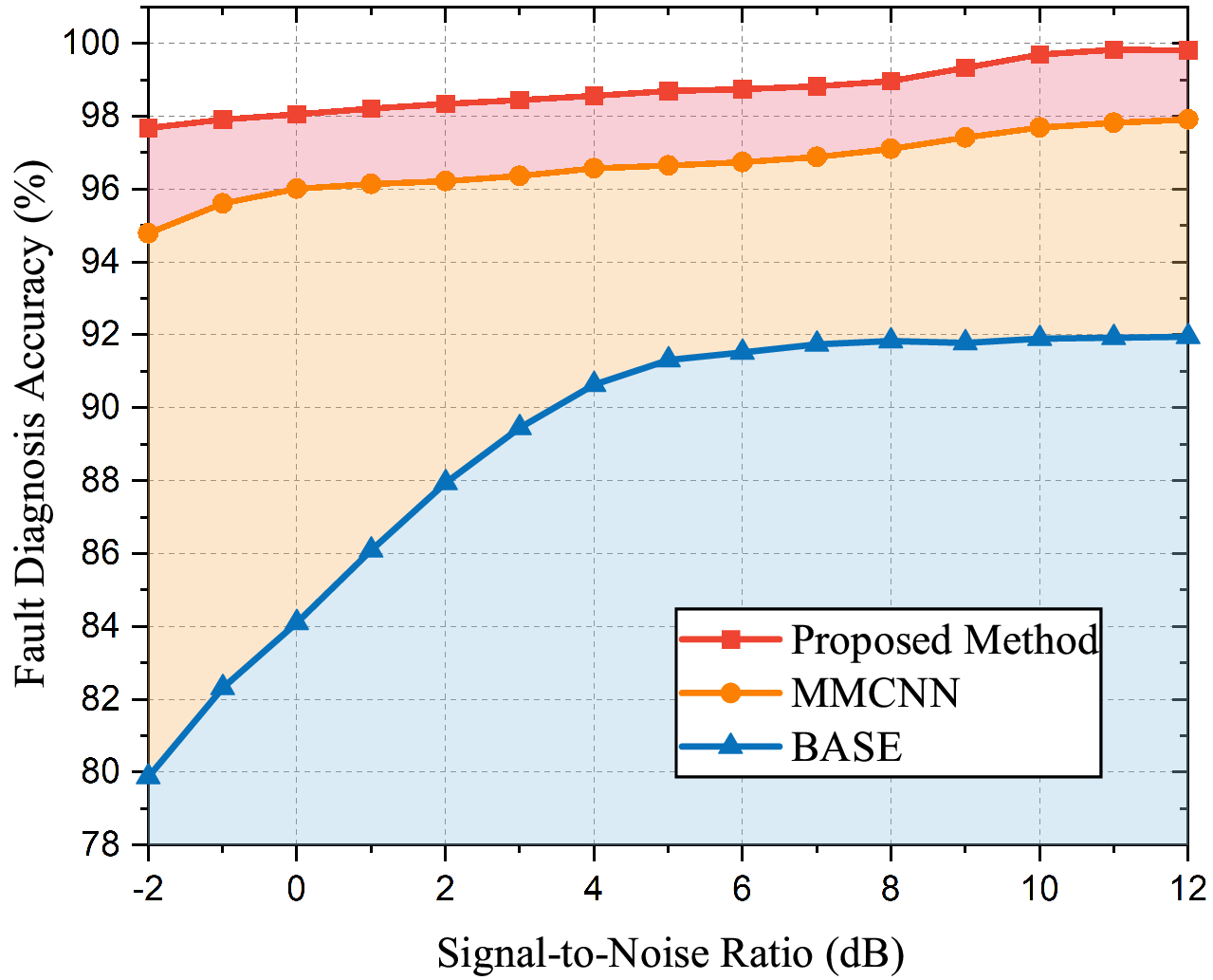}
  \caption{The experimental results of the robustness (SNR) comparison under noisy environments.}
  \label{snrs}
\end{figure}

In industrial scenarios, complicated bearing systems are coordinated by multiple workpieces. These workpieces with different characteristics individually generate the interferential signals that would harm the accuracy of trained fault diagnosis models. Therefore, we emphasize the importance of the model's robustness under noisy work conditions. In real scenarios, the interference is mostly a combination of noise from various probability distributions. According to the central limit theorem, the interference will be normally distributed if the population of its constituent is sufficiently large. Thus, for the complicated bearing systems, it would be a simple but efficient simulation to use Gaussian noise as the experimental interference. Here, the Signal-to-Noise Ratio (SNR) is adopted to indicate the interference level, which is as follows.
\begin{align*}
  & \text{SNR} \left(\text{dB}\right) = 10 \log_{10}\left(P_{\text{signal}} / P_{\text{noise}}\right)
\end{align*}
where $P_{\text{signal}}$ and $P_{\text{noise}}$ are the power of the raw signal and additive Gaussian noise, respectively. For comparison, we used our pure baseline, BASE, and another method, MMCNN, proposed by Ma et.al\citep{MMCNN}. Our baseline adopts no noise-robust methods while MMCNN uses multimodal time-frequency images to enhance the robustness. In this experiment, we evaluate CDHM, BASE, and MMCNN under SNRs ranging from -2dB to 12 dB on the CWRU dataset. The corresponding results are shown in Fig.\ref{snrs}.

The SNRs lower than 0dB denote the proportion of the noise is much larger. We observe that when SNRs are lower than 6dB, CDHM begins to outperform BASE significantly. Regardless of how SNR changes, CDHM can maintain a leading advantage relative to MMCNN. Such results reflect the surpassing robustness of CDHM under noisy environments. From another perspective, the improvement of robustness literally demonstrates that benefitted from the causal disentanglement, CDHM can eliminate the fault-irrelevant interference.

\section{Conclusion}
In this paper, a novel fault diagnosis method, CDHM, is proposed. CDHM simultaneously tackles precise fault diagnosis and accurate cross-domain diagnosis through the causal learning network and transfer learning network, respectively. In the causal learning network, the embedded causal mechanism of bearing failures is structured into a DAG, and a sequential network is thus established to learn these causal relations. Subsequently, a precise and robust representation is separated from the input signal. We reformulated ELBO to optimize this network. For transfer learning tasks, domain divergence is initially handled by the ability to learn the native causal mechanism in bearing failures, while the learning process of discrimination loss enhanced this ability. Then, the unsupervised MMD loss is adopted to narrow the gap between different domains and thus, the disentangling faculty is transferred to other domains. The experimental results demonstrate CDHM can achieve higher accuracy and also exhibit stronger robustness for fault detection tasks under cross-domain and interference conditions. In the future, to facilitate the application of fault diagnosis algorithms, we will further explore the causal mechanism mined in the complicated multi-fault task as well as the open-set fault diagnosis task.

\section*{Declarations}

\begin{itemize}
\item \textbf{Funding} The research leading to these results received funding from the National Key Research and Development Program of China under Grant Agreement No.2020YFB1711700.
\item \textbf{Competing interests} The authors have no competing interests to declare that are relevant to the content of this article.
\item \textbf{Data availability} The data that support the findings of this study are available from the corresponding author upon request.
\end{itemize}

\bibliography{sn-bibliography}

\end{document}